\documentclass[letterpaper, 10 pt,journal, twoside]{ieeetran}  

\IEEEoverridecommandlockouts                              

\usepackage{graphics} 
\usepackage{epsfig} 
\usepackage{times} 
\usepackage{amsmath} 
\usepackage{amssymb}  
\usepackage{amsfonts}
\usepackage[colorlinks,linkcolor=black,anchorcolor=black,citecolor=black,urlcolor=black]{hyperref}
\usepackage[table]{xcolor}
\usepackage{booktabs}
\usepackage{multirow}
\usepackage{graphicx}
\usepackage{subfigure}
\usepackage{cite}
\usepackage{float}
\usepackage{makecell}

\usepackage{etoolbox}
\makeatletter
\patchcmd{\@makecaption}
  {\scshape}
  {}
  {}
  {}
\makeatletter
\patchcmd{\@makecaption}
  {\\}
  {.\ }
  {}
  {}
\makeatother
\def\tablename{Table}

\graphicspath{{img/}}

\title{MacFormer: Map-Agent Coupled Transformer \\for Real-time and Robust Trajectory Prediction}

\author{Chen Feng$^{2}$, Hangning Zhou$^{3}$, Huadong Lin$^{3}$, Zhigang Zhang$^{3}$, \\ Ziyao Xu$^{3}$, Chi Zhang$^{3}$, Boyu Zhou$^{1,\dag}$,
 and Shaojie Shen$^{2}$
\thanks{Manuscript received: April 24, 2023; Revised: July 24, 2023; Accepted: August 7, 2023.
This paper was recommended for publication by Editor Jens Kober upon evaluation of the Associate Editor and Reviewers' comments.
This work was supported by The Research Grants Council General Research Fund (RGC GRF) project RMGS20EG20.}
\thanks{$^{1}$School of Artificial Intelligence, Sun Yat-Sen University, Zhuhai, China.}
\thanks{$^{2}$Department of Electronic and Computer Engineering, The Hong Kong University of Science and Technology, Hong Kong, China.}
\thanks{$^{3}$Megvii Research, Beijing, China.}%
\thanks{Email: {\tt\footnotesize cfengag@ust.hk}, {\tt\footnotesize zhouby23@mail.sysu.edu.cn}}
\thanks{\textbf{$^{\dag}$ Corresponding Author}}
\thanks{Digital Object Identifier (DOI): see top of this page.}}

\providecommand{\keywords}[1]{\textbf{\textit{Index terms---}} #1}
\begin{document}

\maketitle
\markboth{IEEE Robotics and Automation Letters. Preprint Version. Accepted August, 2023}{Feng \MakeLowercase{\textit{et al.}}: MacFormer: Map-Agent Coupled Transformer for Real-time and Robust Trajectory Prediction}

\begin{abstract}

Predicting the future behavior of agents is a fundamental task in autonomous vehicle domains.
Accurate prediction relies on comprehending the surrounding map, which significantly regularizes agent behaviors. 
However, existing methods have limitations in exploiting the map and exhibit a strong dependence on historical trajectories, which yield unsatisfactory prediction performance and robustness.
Additionally, their heavy network architectures impede real-time applications.
To tackle these problems, we propose Map-Agent Coupled Transformer (MacFormer) for real-time and robust trajectory prediction. 
Our framework explicitly incorporates map constraints into the network via two carefully designed modules named coupled map and reference extractor.
A novel multi-task optimization strategy (MTOS) is presented to enhance learning of topology and rule constraints.
We also devise bilateral query scheme in context fusion for a more efficient and lightweight network.
We evaluated our approach on Argoverse 1, Argoverse 2, and nuScenes real-world benchmarks, where it all achieved state-of-the-art performance with the lowest inference latency and smallest model size.
Experiments also demonstrate that our framework is resilient to imperfect tracklet inputs.
Furthermore, we show that by combining with our proposed strategies, classical models outperform their baselines, further validating the versatility of our framework.

\end{abstract}

\begin{keywords}
    \textbf{Deep Learning Methods; Representation Learning; Autonomous Vehicle Navigation}
\end{keywords}

\section{Introduction}
\label{sec:intro}

\IEEEPARstart{A}{ccurate} trajectory prediction of nearby agents is crucial for safe navigation of autonomous vehicles. 
Understanding the surrounding map is essential to enhance prediction performance, as it imposes topology and rule constraints greatly regularizing agent behaviors. 
Specifically, the topology constraint requires consistency with lane types, while the rule constraint prohibits driving outside drivable areas and prefers trajectories near centerlines. 

\begin{figure}[t]
    \begin{center}
        \includegraphics[width=0.8\columnwidth]{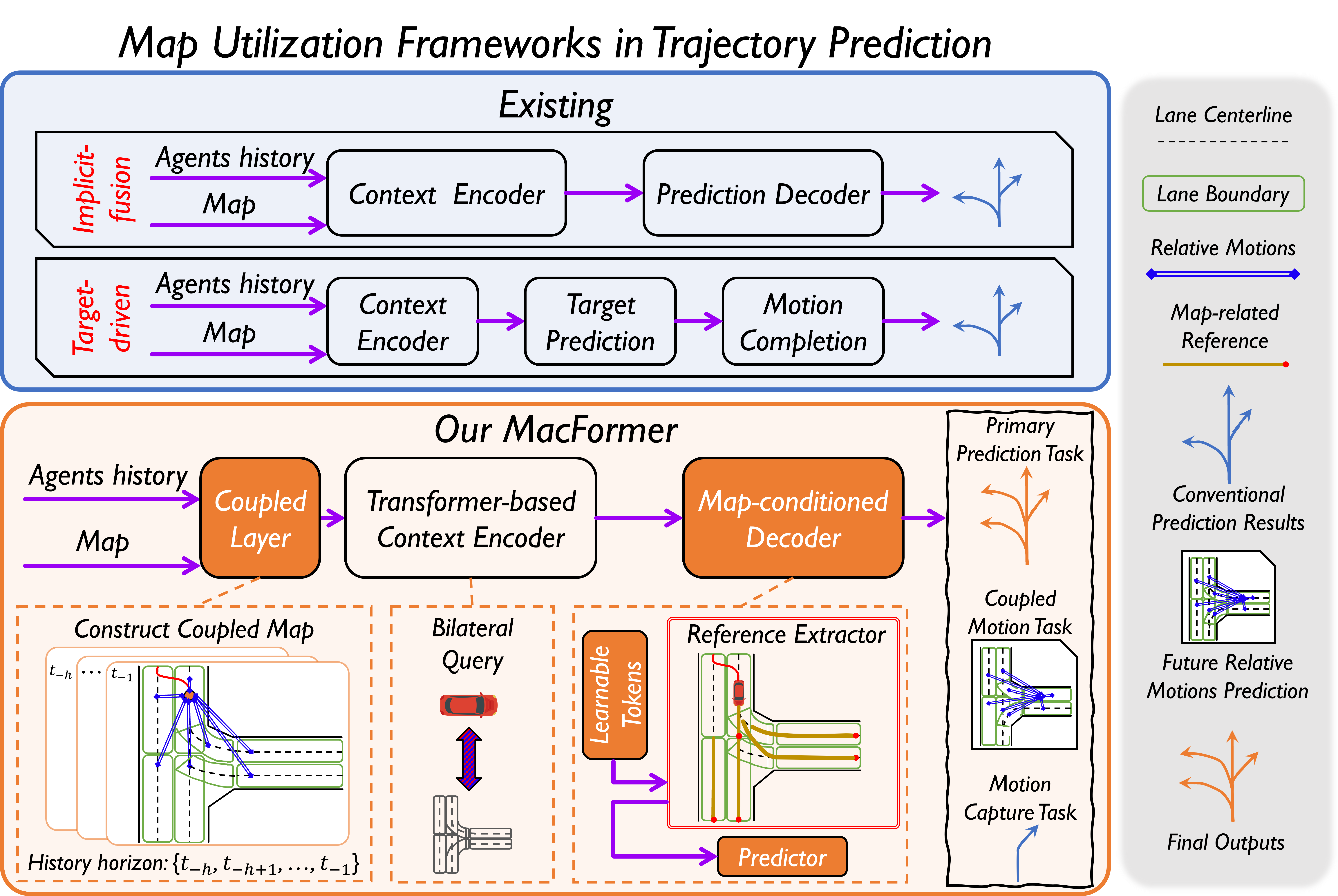}   
        \vspace{-0.5cm}
    \end{center}
    \caption{\label{fig:top_fig} \textbf{The overview of map utilization in trajectory prediction.} Existing works (Top) take \textit{implicit-fusion} or \textit{target-driven} manner, which are yet to exhaust the map. 
    In contrast, our method (Bottom) explicitly and sufficiently leverage map constraints, effectively capturing prediction uncertainty and enhancing robustness to imperfect tracklets. (Sect.\ref{sec:intro})}
    \vspace{-0.6cm}
\end{figure}

To integrate map constraints into models, existing works utilize the map mainly in two manners: \textit{implicit-fusion} and \textit{target-driven}. 
The \textit{implicit-fusion} models \cite{gao2020vectornet, konev2021motioncnn, liang2020learning, ngiam2022scene, varadarajan2022multipath++, liu2021multimodal, ye2021tpcn, zhang2022trajectory,zhou2022hivt, shi2022motion} use implicit frameworks to fuse the historical motions of maps and agents into high-level latent features, which are then decoded to predict multiple trajectories. 
However, to achieve high-level latent features, a deeper network design is necessary, resulting in a heavier architecture. 
Specifically, multiple layers are typically stacked in the context fusion stage to effectively fuse map information and agent motions.
Meanwhile, during training, these models only minimize errors between ground-truth and predicted trajectories without explicitly using map constraints. 
This results in the model output being dominated by past motions and ignoring map information due to high correlation between historical and future movements \cite{makansi2021you}. 
Thus, these models are likely to produce inadequate probabilistic predictions or infeasible trajectories, \textit{e.g.} causing mode collapse or driving outside the drivable area.

On the other hand, \textit{target-driven} models \cite{zhao2020tnt, gu2021densetnt, gilles2021home, gilles2021gohome, gilles2021thomas,wang2022ganet} reduce prediction uncertainty by selecting plausible endpoint targets from the map and using them to complete future motions. 
Nevertheless, these models only utilize map information when choosing endpoints, and capturing future behaviors solely based on a target remains challenging. 
Hence, there is still significant prediction uncertainty. 
Additionally, these models suffer from a performance-cost trade-off, \textit{i.e.}, fewer target candidates lead to performance decline while more candidates result in extensive computation cost. 
In a nutshell, current methods lack proper use of maps and rely heavily on historical trajectories, resulting in unsatisfactory prediction performance when faced with imperfect historical motions. 
Moreover, they employ heavy network architectures, particularly in context fusion, which may hinder real-time capability in practical applications.

\begin{figure*}[t]
	\begin{center}
      \includegraphics[width=1.99\columnwidth]{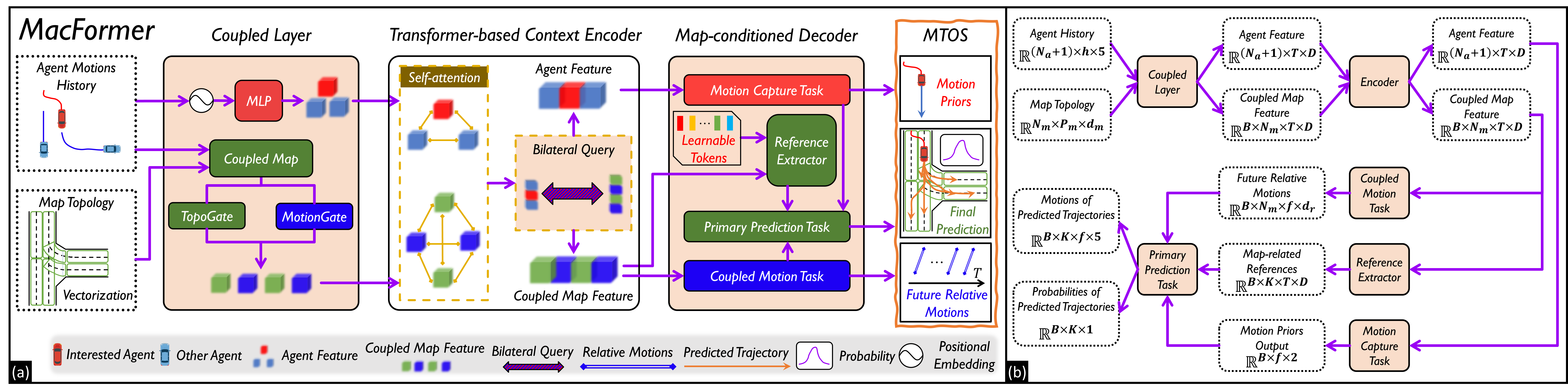}     
      \vspace{-0.5cm}
	\end{center}
   \caption{\label{fig:sys} (a) The system overview of \textbf{MacFormer}. (b) The detailed implementation of \textbf{MacFormer} and output size of each operation.}
   \vspace{-0.6cm}
\end{figure*}

To address the above-mentioned issues, we propose \textbf{MacFormer}, an one-stage \textbf{M}ap-\textbf{A}gent \textbf{C}oupled Trans\textbf{Former}, which directly couples map constraints with agent motions for real-time and robust trajectory prediction (Fig.\ref*{fig:top_fig}). 
In order to enable direct utilization of the coupled relation between the map and agent, two modules named \textbf{coupled map} and \textbf{reference extractor} are carefully designed, which explicitly integrate map constraints into the system. 
Furthermore, to assure that the map constraints are well learned during training, we propose a multi-task optimization strategy (\textbf{MTOS}). 
It ensures map-constrained prediction by forecasting future relative motions in coupled map, capturing motion priors, and outputting final predictions with corresponding probabilities via map-conditioned regression. 
Lastly, to enhance computational efficiency and promote real-time prediction, a concise \textbf{bilateral query} scheme for context fusion is devised. 
It allows parallel cross-domain information interaction between the map and agent from their distinct perspectives and shared usage of their affinity matrix, significantly reducing the time and space complexity of context fusion.

The proposed method was evaluated on three large-scale real-world benchmarks (Argoverse 1\&2 \cite{chang2019argoverse, wilson2021argoverse} and nuScenes \cite{caesar2020nuscenes}) 
and achieved state-of-the-art performance with significantly lower inference latency and fewer parameters. 
We also conducted extensive experiments to validate the effectiveness and robustness of our framework. 
Furthermore, we introduced our framework to classical prediction models, enhancing their performance while reducing parameters.
In summary, the contributions of this paper are:

1) A map-agent coupled framework for real-time and robust trajectory prediction, which efficiently and effectively integrates map constraints via coupled map, reference extractor, and multi-task optimization strategy (MTOS).

2) An efficient and lightweight context fusion scheme, bilateral query, which allows parallel context fusion between map and agent from their individual views.

3) Extensive evaluation in multiple real-world benchmarks. It shows that \textbf{MacFormer} achieves state-of-the-art performance, while it is faster and more lightweight than existing models. 
Experiments also demonstrate the versatility and robustness of the proposed framework, which enhances classical prediction methods while maintaining satisfactory resilience to imperfect upstream results.

\section{Related Work}
\label{sec:related_work}

\noindent\textbf{Map utilization.} Map utilization has received much attention recently, where existing approaches mainly fall into two different directions: \textit{implicit-fusion} and \textit{target-driven}.

1) \textbf{\textit{Implicit-fusion.}} \textit{Implicit-fusion} models use different backbones to encode map and fuse it with agent motions in high-level latent feature. 
By optimizing prediction outputs, they aim to capture the map structure in such fusion. 
Early works \cite{konev2021motioncnn, park2020diverse} used CNNs to encode the map and agents together into an image, then predicted trajectories using a fully connected layer. 
However, this approach cannot model relations between different scenario elements effectively. 
VectorNet \cite{gao2020vectornet}, LaneGCN \cite{liang2020learning}, and DSP \cite{zhang2022trajectory} all vectorized the map and adopted graph-based architectures to extract features and interactions of agents and map instead. 
Besides, TPCN \cite{ye2021tpcn} migrated point cloud learning backbone to implement the feature fusion. 
To cover long-range prediction, mmTransformer \cite{liu2021multimodal}, SceneTransformer \cite{ngiam2022scene} and MultiPath++ \cite{varadarajan2022multipath++} leveraged the global receptive field of Transformer to capture whole context fusion. 
Nevertheless, \textit{implicit-fusion} still leads to mode collapse or infeasible forecasts that violate map constraints, which illustrates their unsatisfactory map utilization and prediction stability.

2) \textbf{\textit{Target-driven.}} To overcome the limitations in \textit{implicit-fusion} models, TNT \cite{zhao2020tnt} and DenseTNT \cite{gu2021densetnt} decomposed prediction objectives into two stage: target prediction and motion completion. 
They first sampled several target candidates for the agent from map, selected qualified targets from candidates, then completed full trajectory conditioned on each target. 
Similarly, HOME \cite{gilles2021home}, GOHOME \cite{gilles2021gohome} and THOMAS \cite{gilles2021thomas} forecasted a heatmap and greedily sampled target candidates to implement target prediction before generating full trajectories. 
However, despite the high intent correlation of targets, such models still cannot effectively capture motions of the entire process, while also incurring redundant costs due to large quantities of unselected candidate.

In contrast, our method tackles these problems by directly and sufficiently leveraging map, which utilizes the coupled relation by coupled map, 
enables efficient map-constrained predictions via reference extractor, considers map constraints during optimization via MTOS.

\noindent\textbf{Transformer.} Transformer \cite{vaswani2017attention} has been successfully applied in various fields, \textit{e.g.}, natural language processing and computer vision. 
Recently, some works \cite{ngiam2022scene, liu2021multimodal, varadarajan2022multipath++} have utilized it and its variants for trajectory prediction due to its capability of global and flexible feature extraction that are highly applicable for context interaction. 
To leverage this architecture, our proposed method devises a Transformer-based context encoder and proposes an effective query module in reference extractor to extract map-related references from coupled map feature unlike a concurrent work MTR \cite{shi2022motion} that designs a motion query pair for learning motion modes in decoder.

\begin{figure}[t]
	\begin{center}
      \vspace{-0.2cm}
      \includegraphics[width=0.90\columnwidth]{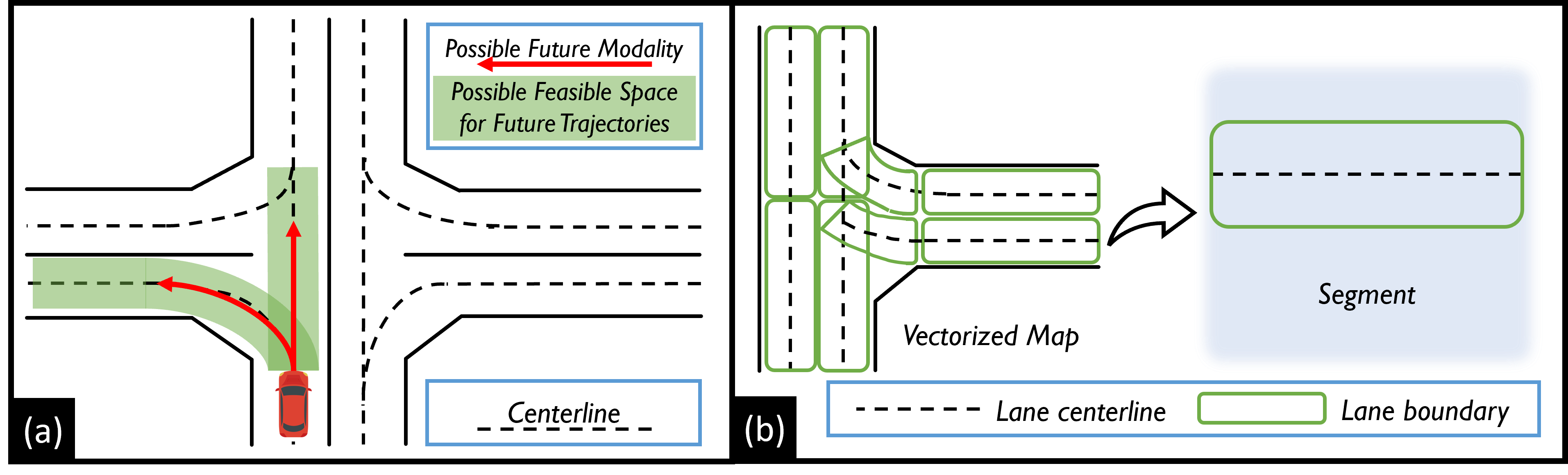}     
      \vspace{-0.5cm}
	\end{center}
   \caption{\label{fig:mmcons_seg} (a) Illustration of map constraints on multi-modality. (b) Illustration of segments in vectorized map.}
   \vspace{-0.5cm}
\end{figure}

\section{Problem Formulation}
\label{sec:problem_formulation}

Given the past movements of the interested agent ${\rm \textbf{s}}_{H}$ over $h$ timestamps, 
the nearby map topology $\mathcal{M}$ (position, type, and connectivity of lanes), and observations of $N_{a}$ other agents $\mathcal{S}=\{{\rm \textbf{s}}_{1}, ..., {\rm \textbf{s}}_{N_{a}} \}$, 
existing works aim to determine the distribution $P(\mathcal{Y}|{\rm \textbf{s}}_H,\mathcal{S},\mathcal{M})$ for future motions represented by $\mathcal{Y}=\{y_0,...,y_{f-1}\}$ spanning $f$ timestamps.
The entire time range is denoted as $T=h+f$ timestamps.

However, as mentioned before, the formulation of existing approaches cannot sufficiently exploit map information.
Thus, we propose MTOS to effectively learn map constraints (Sect.\ref{sub:mtos}). 
In our formulation, the optimization objective also includes predicting future relative motions $\mathcal{R}$ and motion priors $\mathcal{J}$.
Specifically, $\mathcal{M}$ is divided into several segments, each of which is discretized as a set of points.
Each discrete point contains its position, lane type, and connected segments.  
$\mathcal{R}$ denotes the vectors pointing from the closest point in each segment towards ${\rm \textbf{s}}_{H}$ in the future horizon.
$\mathcal{J}$ is represented as future positions of ${\rm \textbf{s}}_{H}$ inferred merely from historical trajectory to capture the scale and pattern of history.
Hence, based on our coupled map $\mathcal{C}_{\mathcal{M}}({\rm \textbf{s}}_{H}, \mathcal{M})$ that includes historical relative motions and map topology, 
the objective of \textbf{MacFormer} is formulated as a joint distribution $P(\mathcal{Y}, \mathcal{R}, \mathcal{J}|\mathcal{X})$, 
where we denote $\mathcal{X}=({\rm \textbf{s}}_{H}, \mathcal{S},\mathcal{C}_{\mathcal{M}}({\rm \textbf{s}}_{H}, \mathcal{M}))$. 
Due to the conditional independence of $\mathcal{R}$ and $\mathcal{J}$ conditioned on $\mathcal{X}$, we factorize it as two marginal distributions and a conditional distribution to diminish the complexity of learning a joint distribution:
\begin{equation}
    P(\mathcal{Y}, \mathcal{R}, \mathcal{J}|\mathcal{X}) = P(\mathcal{Y}|\mathcal{R}, \mathcal{J}, \mathcal{X})P(\mathcal{R}|\mathcal{X})P(\mathcal{J}|\mathcal{X}).
    \label{eq:1}
\end{equation}

In fact, $P(\mathcal{Y}|\mathcal{R}, \mathcal{J}, \mathcal{X})$ is still not tractable for the model due to its high uncertainty and complicated multi-modality. 
As illustrated in Fig.\ref{fig:mmcons_seg}(a), trajectories are typically confined within a space around possible centerlines. 
Under this observation, the entire difficult prediction problem can be decomposed into two easier tasks: 1) predicting the possible centerlines, named map-related references $\Psi=\{\psi_1,...\psi_K\}$ where $\psi$ denotes one possible map-related reference.
and 2) predicting trajectories conditioned on each of the map-related references.
Hence, we decompose $P(\mathcal{Y}|\mathcal{R}, \mathcal{J}, \mathcal{X})$ according to law of total probability to alleviate the high uncertainty.
We have $P(\psi|\mathcal{R},\mathcal{J}, \mathcal{X})$=$P(\psi|\mathcal{X})$ due to the conditional irrelevance of $\psi$ conditioned on $\mathcal{R}$ and $\mathcal{J}$.
As demonstrated in eq.\ref{eq2a}, we firstly extract each plausible map-related reference 
according to agent history and map topology, represented as $P(\psi|\mathcal{X})$ (Sect.\ref{sub:ref_ext}). 
Then, each future trajectory is predicted given the corresponding reference as the guidance, \textit{i.e.}, $P(\mathcal{Y}|\psi,\mathcal{R},\mathcal{J},\mathcal{X})$ (Sect.\ref{sub:mtos}).
Therefore, we expect all possible $\psi$ can capture plausible and complicated multi-modality to better facilitate trajectory prediction.
This decomposition aims to narrow the prediction space, enabling each prediction near its corresponding reference.
\begin{align}
\label{eq2a}
& P(\mathcal{Y}|\mathcal{R}, \mathcal{J}, \mathcal{X})
= \sum\limits_{\psi\in\Psi} P(\mathcal{Y}|\psi,\mathcal{R},\mathcal{J},\mathcal{X})P(\psi|\mathcal{R},\mathcal{J}, \mathcal{X}), \nonumber \\
& P(\mathcal{Y}|\mathcal{R}, \mathcal{J}, \mathcal{X})
= \sum\limits_{\psi\in\Psi} P(\mathcal{Y}|\psi,\mathcal{R},\mathcal{J},\mathcal{X})P(\psi|\mathcal{X}).
\end{align}

Furthermore, our framework can be extended to joint prediction for all agents in the scenario.
The objective is predicting $\mathcal{Y}, \mathcal{R}, \mathcal{J}$ of each agent conditioned on its corresponding 
coupled map $\mathcal{C}_\mathcal{M}$ and history of all $N_a$+$1$ agents.
For all agents, all parameters in our model are shared and all operations are parallel.
Thus, we define the number of predicted agents as $B$, 
$B$=$1$ for single prediction while $B$=$N_a$+$1$ for joint prediction (practical usage for autonomous navigation system).
For brevity, we set $B$=$1$ to clearly introduce our method.

\section{System Overview}
\label{sec:system_overview}

The proposed framework, as shown in Fig.\ref{fig:sys}, consists of three main components. 
Firstly, coupled layer extracts historical motion features and constructs coupled map.
It calculates historical relative motions between the map and agent in each timestamp.
Then, coupled map is defined as the combination of historical relative motions and map topology (Sect.\ref{sub:coupled_layer}).
Subsequently, MotionGate and TopoGate are devised to respectively extract its temporal and spatial features.
Secondly, Transformer-based context encoder employs self-attention for social interaction within respective domain (map or agent), then efficiently achieves parallel context fusion between them using our bilateral query scheme (Sect.\ref{sub:encoder}).
Thirdly, map-conditioned decoder learns corresponding map-related references from coupled map via reference extractor to guide predicted trajectories spatially along some centerlines or their combination (Sect.\ref{sub:ref_ext}). 
To ensure predictions are constrained by maps, we develop multi-task optimization strategy (MTOS).
Coupled motion task forecasts future relative motions in coupled map directly coupling the predictions with the map and imposing map constraints on trajectories.
Motion capture task conventionally predicts one future trajectory to capture motion priors by learning scale and pattern of history.
Lastly, primary prediction task regresses final predictions with corresponding probabilities conditioned on map-related references and outputs from above two tasks (Sect.\ref{sub:mtos}).

\section{Methodology}
\label{sec:methodology}

\subsection{Coupled Layer}
\label{sub:coupled_layer}
To begin, we apply the agent-centric strategy to normalize all inputs based on the coordinate system of the interested agent. 
Then, coupled layer processes them for extracting motion features of agents and constructing coupled map.

In our formulation, the motions of agents $\rm \textbf{s}$ are represented as the set of locations ($x,y$), azimuth angles ($\alpha$), and velocity ($v$) $[x,y,\cos\alpha,\sin\alpha,v]$. 
For the future horizon, we use a mask on these timestamps then concatenate it with agents' historical motions (${\rm \textbf{s}}_{H}$ and $\mathcal{S}$). 
Subsequently, positional embedding ${\rm PE}(\cdot)$ is applied on each timestamp to reinforce the identification of each timestamp \cite{vaswani2017attention}. 
Afterwards, we deploy a multi-layer perceptron network ${\rm MLP}(\cdot)$ for motions inputs. The agent feature in coupled layer $\mathcal{F}_{A}^{CL}$ follows:
\begin{equation}
    \mathcal{F}_{A}^{CL} = {\rm MLP}({\rm PE}([{\rm \textbf{s}}_{H}, \mathcal{S}]))\in \mathbb{R}^{(N_{a}+1) \times T \times D}.
\end{equation}
with $D$ the feature dimension and $[\cdot]$ the concatenation.

\begin{figure}[t]
	\begin{center}
      \includegraphics[width=0.85\columnwidth]{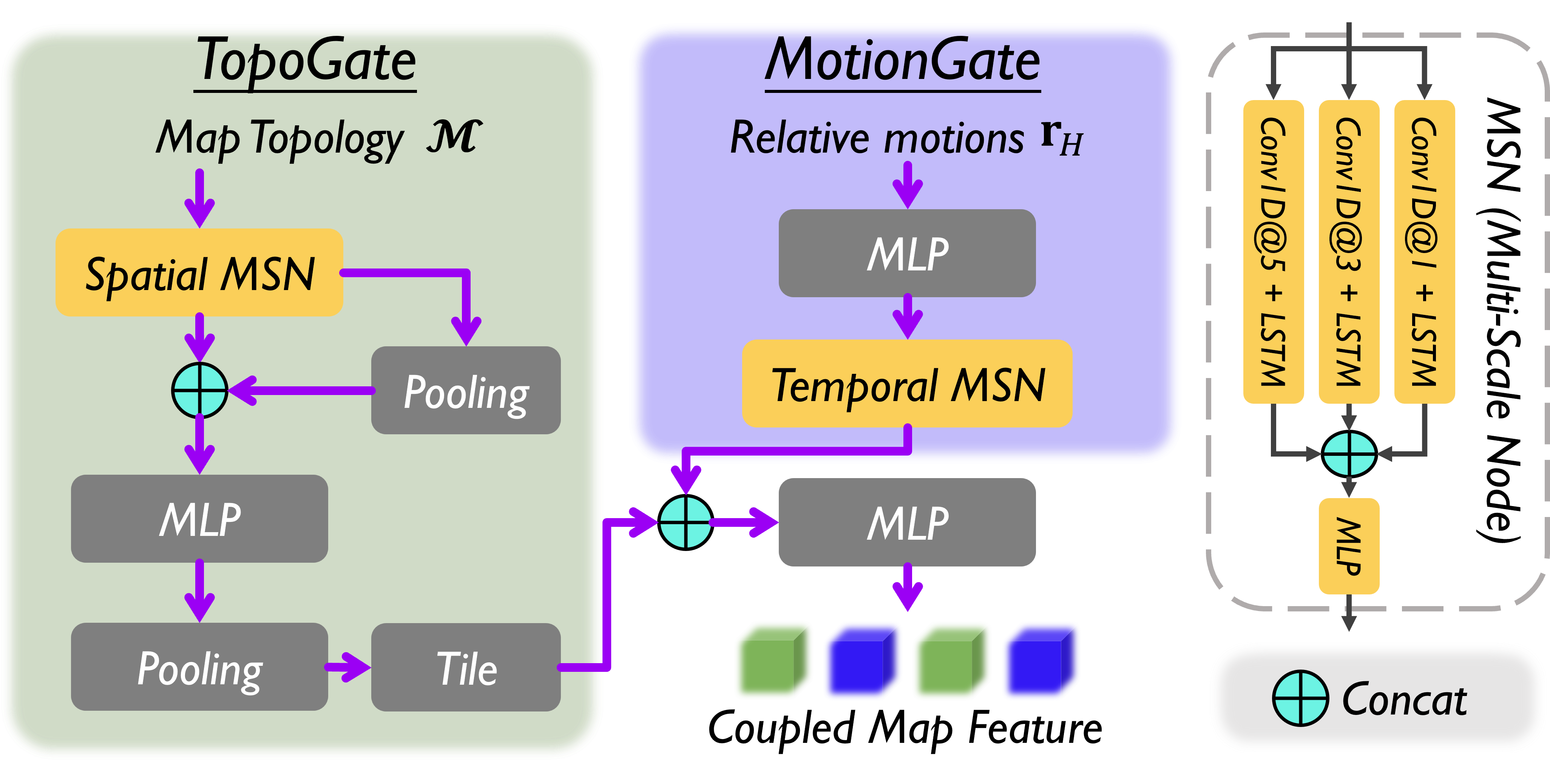}     
      \vspace{-0.5cm}
	\end{center}
   \caption{\label{fig:gate} The detailed structure of \textbf{TopoGate} and \textbf{MotionGate}. (Sect.\ref{sub:coupled_layer})}
   \vspace{-0.5cm}
\end{figure}

\noindent\textbf{Construct Coupled Map.} We have chosen the vectorized map to describe map topology $\mathcal{M}$ due to its compactness and completeness of map representation \cite{gao2020vectornet}.
As shown in Fig.\ref{fig:mmcons_seg}(b), the lane area is divided into $N_{m}$ segments denoted as $\mathcal{M}=\{m_{i}|i\in[1,...,N_{m}]\}$, where each segment is discretized into a set of $P_{m}$ points. 
Each point contains $d_{m}$ attributes, \textit{e.g.}, location, predecessor and successor points, road type, connected segments. 
Then, we calculate the historical relative motions ${\rm \textbf{r}}_{H}$ between ${\rm \textbf{s}}_{H}$ and $\mathcal{M}$. 
Here, ${\rm \textbf{r}}_{H}$ refers to the vectors from the closest point in each segment to the interested agent at each timestamp, 
including $d_r$ attributes (distance and direction).
The relative motions calculation function $\varphi(\cdot)$ follows:
\begin{align}
    &\varphi({\rm \textbf{s}}_{H},\mathcal{M}) = \{ {\rm \textbf{s}}_{H}-m_i(l)|l=
    \mathop{\arg\min}\limits_{p}||{\rm \textbf{s}}_{H}-m_i(p)||_2, \nonumber\\
    &p\in[1,...,P_m], i\in[1,...,N_m]\}.
\end{align}
where $m_i(p)$ is one discrete point of the segment $m_i$.
Like the operations performed on agents, the relative motions ${\rm \textbf{r}}_{H}$ are also subject to a future horizon mask and positional embedding at each timestamp.
The coupled map $\mathcal{C}_{\mathcal{M}}$ is then formed by combining $\mathcal{M}$ and ${\rm \textbf{r}}_{H}$, as illustrated below:
\begin{equation}
\begin{split}
    &{\rm \textbf{r}}_{H}={\rm PE}(\varphi({\rm \textbf{s}}_{H},\mathcal{M}))\in \mathbb{R}^{B \times N_{m} \times T \times d_{r}},\\
    &\mathcal{C}_{\mathcal{M}}({\rm \textbf{s}}_{H}, \mathcal{M})=\{{\rm \textbf{r}}_{H}, \mathcal{M}\}.
\end{split}
\end{equation}

The topology constraint requires that agent behaviors align with the lane types, namely leftmost, middle, and rightmost lanes. 
For instance, if agents are in the leftmost lane at an intersection, they can only turn left or proceed straight ahead because there is no lane connected to the leftmost one that allows right turns on the map topology.
Thus, as depicted in Fig.\ref{fig:gate}, we devise TopoGate $\mathcal{G}^{T}$ to effectively extract map topology where the core operation is the Multi-Scale Node (MSN).
Specifically, MSN contains three 1D Conv layers with different kernel sizes followed by LSTM to facilitate multi-scale and sequential feature extraction.
TopoGate adopts a spatial MSN with all operations on $P_{m}$ (points) dimension for capturing spatial topology.
We then tile the map topology feature by $T$-fold consistent with the shape ${\rm \textbf{r}}_{H}$ as $\mathcal{G}^{T}(\mathcal{M})\in\mathbb{R}^{B \times N_{m}\times T \times D}$.
Similarly, MotionGate $\mathcal{G}^{M}$ is proposed for leveraging the time sequence of historical relative motions ${\rm \textbf{r}}_{H}$ as shown in Fig.\ref{fig:gate}.
First, we deploy an MLP block for relative motion encoding.
Then, temporal MSN is applied on $T$ (time) dimension of ${\rm \textbf{r}}_{H}$ to extract temporal motion information resulting in $\mathcal{G}^{M}({\rm \textbf{r}}_{H})\in\mathbb{R}^{B \times N_{m}\times T \times D}$.
By combining these two gates together, we obtain the coupled map feature in coupled layer $\mathcal{F}_{\mathcal{C}_{\mathcal{M}}}^{CL}$:
\begin{equation}
    \mathcal{F}_{\mathcal{C}_{\mathcal{M}}}^{CL}={\rm MLP}([\mathcal{G}^{T}(\mathcal{M}), \mathcal{G}^{M}({\rm \textbf{r}}_{H})]) \in \mathbb{R}^{B \times N_{m}\times T \times D}.
\end{equation}

\begin{figure}[t]
	\begin{center}
      \includegraphics[width=0.85\columnwidth]{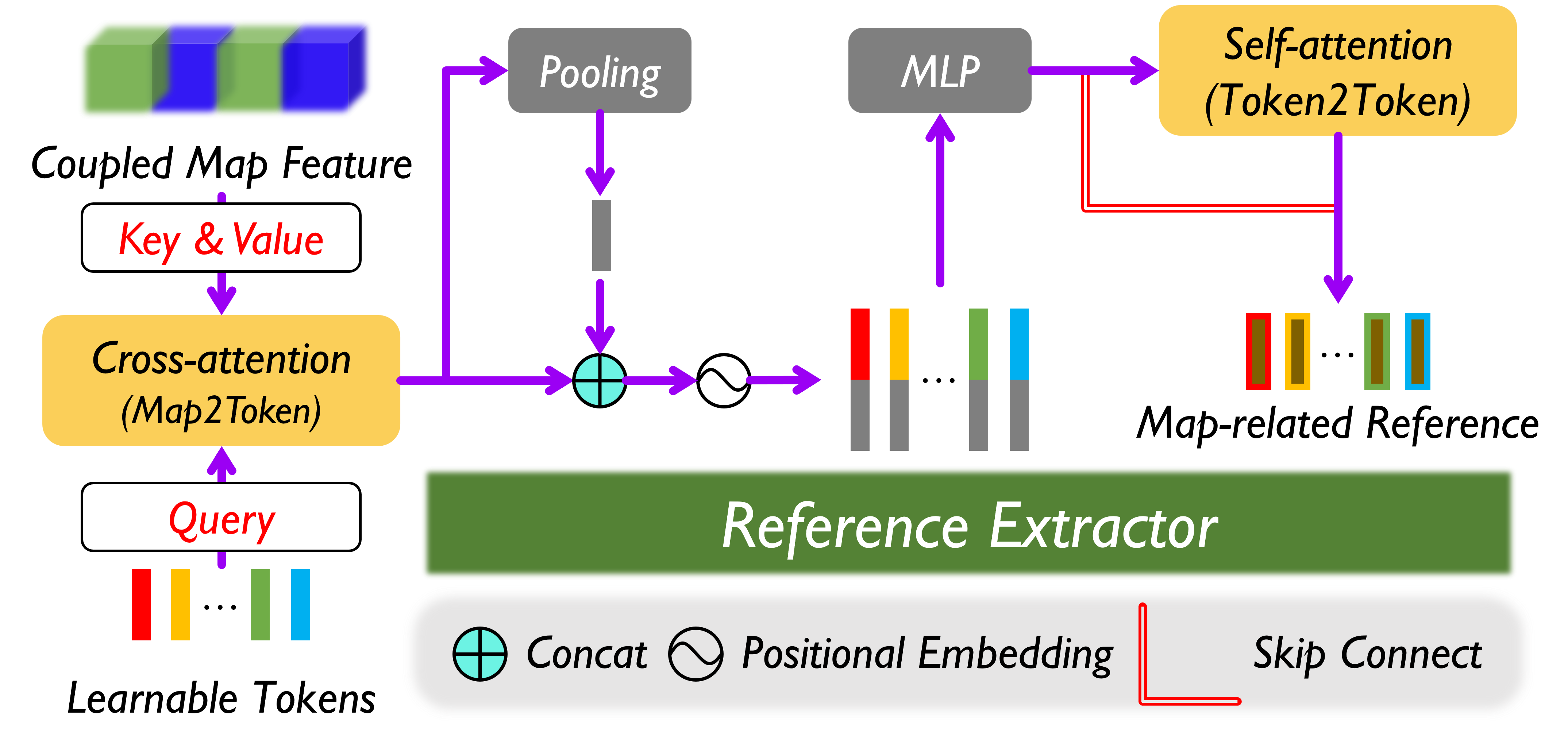}     
      \vspace{-0.5cm}
	\end{center}
   \caption{\label{fig:reference} The overview of the proposed \textbf{reference extractor}. (Sect.\ref{sub:ref_ext})}
   \vspace{-0.5cm}
\end{figure}

\vspace{-0.4cm}
\subsection{Reference Extractor}
\label{sub:ref_ext}

The rule constraint prohibits driving outside the drivable area, and trajectories near centerlines rather than lane boundaries are preferred.
This constraint yields trajectories confined within neighborhood space surrounding the centerlines.
Thus, we propose the reference extractor denoted as $\mathcal{RE}(\cdot)$ that can learn map-related references from the coupled map feature in encoder $\mathcal{F}_{\mathcal{C}_{\mathcal{M}}}^{Encoder}$, as illustrated in Fig.\ref{fig:reference}.
These references guide predictions throughout the entire future process.
This module aims to model $P(\psi|\mathcal{X})$ guiding final $K$ predicted trajectories to capture the entire prediction uncertainty and plausible multi-modality. 
Specifically, we generate $K$ different learnable tokens (each one corresponds to a modality) to extract the map-related reference feature from $\mathcal{F}_{\mathcal{C}_{\mathcal{M}}}^{Encoder}$ by cross-attention. 
Notably, the information learned by tokens includes the spatial topology of map, agent motions, and their coupled relations. 
To ensure the consistency of multi-modality, we concatenate the pooling feature with each token.
Subsequently, positional embedding is applied on each token and self-attention enhances the interaction within multi-modality to avoid mode collapse. 
With the aid of $\mathcal{RE}(\cdot)$, the map-related references feature $\mathcal{F}_{\Psi}$ capturing $P(\Psi|\mathcal{X})$ is given as:
\begin{equation}
    \mathcal{F}_{\Psi}=\mathcal{RE}(\mathcal{F}_{\mathcal{C}_{\mathcal{M}}}^{Encoder})\in \mathbb{R}^{B \times K\times T \times D}.
\end{equation}

\vspace{-0.4cm}
\subsection{Multi-Task Optimization Strategy}
\label{sub:mtos}

We propose a multi-task optimization strategy (MTOS) for effectively incorporating map constraints.
This strategy prevents the network from neglecting map information and enhances its ability to learn map constraints during training.
We discuss each task individually in MTOS, followed by the uniform optimization process.

\noindent\textbf{Coupled Motion Task.} Relative motions effectively bridge the map constraints and trajectories.
If the network can forecast accurate future relative motions $\mathcal{R}$, it demonstrates the network has learned map constraints on agent motions.
Hence, we forecast $\mathcal{R}$ from coupled map feature $\mathcal{F}_{\mathcal{C}_{\mathcal{M}}}^{Encoder}$:
\begin{equation}
    \mathcal{R}={\rm MLP}({\rm SelfAttention}(\mathcal{F}_{\mathcal{C}_{\mathcal{M}}}^{Encoder})) \in \mathbb{R}^{B \times N_{m}\times f \times d_{r}}.
\end{equation}
We pre-label the future states of coupled map $\mathcal{R}_{gt}$ via $\varphi(\cdot)$ defined in Sect.\ref{sub:coupled_layer} as $\mathcal{R}_{gt}=\varphi(\mathcal{Y}_{gt},\mathcal{M})$. $\mathcal{Y}_{gt}$ is the future trajectory ground-truth. 
Thus, the loss term of this task is:
\begin{equation}
\label{loss_couple}
    \mathcal{L}_{couple}=||\mathcal{R}-\mathcal{R}_{gt}||_2^2.
\end{equation}

\begin{figure}[t]
	\begin{center}
      \includegraphics[width=0.99\columnwidth]{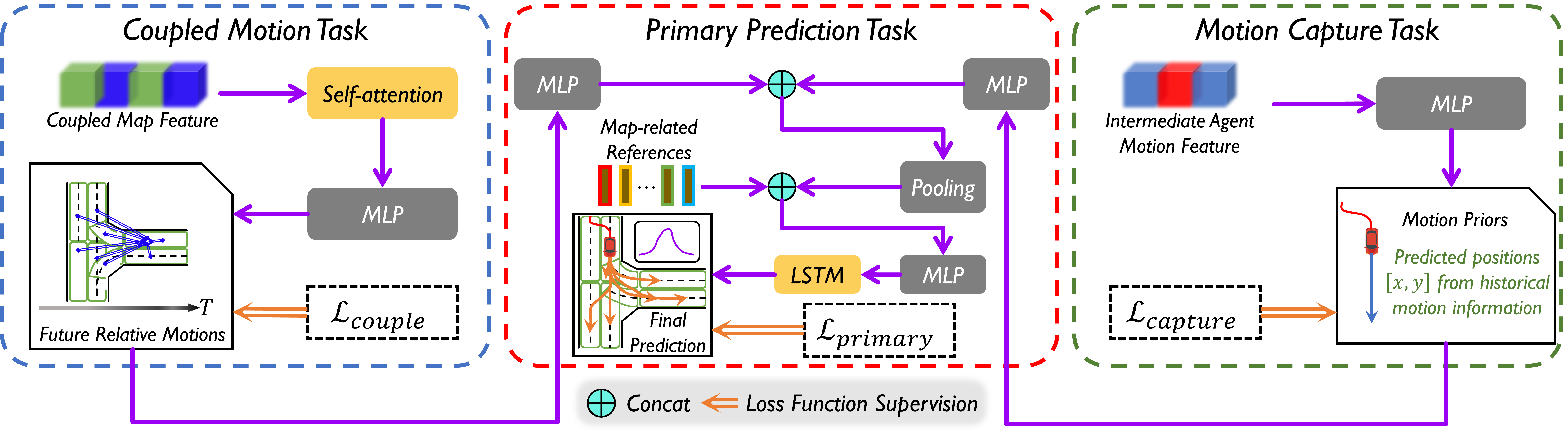}     
      \vspace{-0.8cm}
	\end{center}
   \caption{\label{fig:mc_reg} Illustration of map-conditioned regression. (Sect.\ref{sub:mtos})}
   \vspace{-0.6cm}
\end{figure}

\noindent\textbf{Motion Capture Task.} The objective of this task is to allow the network to capture the motion priors $\mathcal{J}$ of agent inputs \textit{e.g.}, scale and pattern of trajectory.
As stated in Sect.\ref{sec:problem_formulation}, $\mathcal{J}$ is represented as the agent's positions $[x,y]$ inferred only using feature extracted solely from agent motion $\mathcal{F}_{A}^{Encoder}$:
\begin{equation}
    \mathcal{J}={\rm MLP}(\mathcal{F}_{A}^{Encoder}) \in \mathbb{R}^{B \times f \times 2}.
\end{equation}
We denote the errors between $\mathcal{J}$ and positions in $\mathcal{Y}_{gt}$ as $E=\mathcal{J}-\mathcal{Y}_{gt}\{x,y\}$ and the loss function for this task:
\begin{equation}
\label{loss_capture}
    \mathcal{L}_{capture}=\begin{cases}  
        0.5E^2 & \text{if } |E| < 1 \\ 
        |E| - 0.5 & \text{otherwise}.
    \end{cases} 
\end{equation}

\noindent\textbf{Primary Prediction Task.} This task gives the final predictions of our framework. 
According to Sect.\ref{sec:problem_formulation}, we devise map-conditioned regression for explicitly utilizing map constraints.
Specifically, we first concatenate the features of future relative motions $\mathcal{R}$ and motion priors $\mathcal{J}$, denoted as $\mathcal{Z}$.
Then, the motions of predicted trajectories $\mathcal{Y}$ is conditionally regressed based on $\mathcal{Z}$ and map-related references feature $\mathcal{F}_{\Psi}$:
\begin{equation}
    \begin{split}
        &\mathcal{Z} = [{\rm MLP}(\mathcal{R}), {\rm MLP}(\mathcal{J})] \in \mathbb{R}^{B \times N_{m}\times f \times D}, \\
        &\mathcal{Y} = {\rm LSTM}({\rm MLP}([\mathcal{F}_{\Psi}, {\rm Pooling}(\mathcal{Z})])).
    \end{split}
\end{equation}
The regressed motions are $\mathcal{Y} \in \mathbb{R}^{B \times K\times f \times 5}$ with five attributes $[x,y,\cos\alpha,\sin\alpha,v]$ as the same as input.
To ensure smoothness within future trajectories, 
they are decoded via LSTM \cite{hochreiter1997long} which allows prediction of each point is conditioned on its previous predicted points.
Besides, we adopt maximum entropy model to predict the probabilities $P(\mathcal{Y})$ of all the $K$ trajectories as:
\begin{equation}
    P(\mathcal{Y})=\frac{{\rm exp}({\rm MLP}(\mathcal{F}_{\Psi}))}{\sum_{k=1}^{K}{\rm exp}({\rm MLP}({\mathcal{F}_{\Psi}}_{k}))}.
\end{equation}
with $P(\mathcal{Y}) \in \mathbb{R}^{B \times K \times 1}$ and $k$ the indexer of tensor.

To train them with given ground-truth $\mathcal{Y}_{gt}$, we use the Gaussian Mixture Model (GMM) loss for regression and max-margin loss for probability:
\begin{equation}
\begin{split}
    &\mathcal{L}_{gmm}(r,p,r_{gt})=-{\rm log}\sum\limits_{i=1}^{K}p_{i}e^{-\frac{1}{2}||r^{i}-r_{gt}^{i}||_{2}^{2}},\\
    &\mathcal{L}_{m}(p)=\frac{1}{K-1}\sum\limits_{i=1,i\neq\bar{i}}^{K}{\rm max}(0,p_i+\delta-p_{\bar{i}}).
\end{split}
\end{equation}
where $r,p,r_{gt}$ represent motions of predicted trajectories, predicted probability, and trajectory ground-truth accordingly. 
The $p_{\bar{i}}$ is the probability of the prediction closest to the ground-truth.
Additionally, $\delta$ is a margin set at $\frac{1}{K}$. 
Hence, we define the loss term for primary prediction task:
\begin{equation}
\label{loss_primary}
    \mathcal{L}_{primary}=\mathcal{L}_{gmm}(\mathcal{Y},P(\mathcal{Y}),\mathcal{Y}_{gt})+\mathcal{L}_{m}(P(\mathcal{Y})).
\end{equation}

\noindent\textbf{Uniform Optimization Process.} The supervision of the three tasks mentioned above results in a one-stage and fully supervised end-to-end training.
The total loss function, which includes eq.\ref{loss_couple}, \ref{loss_capture}, and \ref{loss_primary}, is formulated as follows:
\begin{equation}
\label{total_loss}
    \mathcal{L}=\mathcal{L}_{primary}+\mathcal{L}_{couple}+\mathcal{L}_{capture}.
\end{equation}
Through minimizing eq.\ref{total_loss}, MTOS achieves the uniform optimization of the objective joint distribution $P(\mathcal{Y}, \mathcal{R}, \mathcal{J}|\mathcal{X})$.

\vspace{-0.3cm}
\subsection{Bilateral Query}
\label{sub:encoder}

The proposed method differs from existing ones in that it incorporates a bilateral query strategy, which efficiently facilitates cross-domain interaction.
Unlike the unilateral query that only supports each agent queries its interested region on the map, our scheme 
allows mutual querying of information between map and agents for more effective context fusion.
Specifically, compared to existing methods, we add an operation that each segment in coupled map feature captures corresponding agent motions within their respective range. 
Given the agent motion feature $\mathcal{F}_{A}^{SI}$ and coupled map feature $\mathcal{F}_{\mathcal{C}_{\mathcal{M}}}^{SI}$ after social interaction, 
we first calculate the affinity matrix ${\rm \textbf{M}}_{aff}$ between agents and coupled map using the learnable linear projection matrix ${\rm \textbf{W}}_{bq}$ and dot-product,
\begin{equation}
\begin{split}
    &{\rm \textbf{z}}^{agent} = {\rm \textbf{W}}_{bq}\mathcal{F}_{A}^{SI}, {\rm \textbf{z}}^{map} = {\rm \textbf{W}}_{bq}\mathcal{F}_{\mathcal{C}_{\mathcal{M}}}^{SI},\\
    &{\rm \textbf{M}}_{aff} = {\rm \textbf{z}}^{agent}{\rm \textbf{z}}^{map} \in \mathbb{R}^{(N_{a}+1)\times N_{m}}.
\end{split}
\end{equation}
Then, we implement the parallel bilateral query process from their respective views based on affinity matrix by the scaled element-product,
\begin{equation}
\begin{split}
    &{\rm \textbf{q}}^{agent} = {\rm \textbf{W}}_{Q}^{agent}\mathcal{F}_{A}^{SI}, {\rm \textbf{q}}^{map} = {\rm \textbf{W}}_{Q}^{map}\mathcal{F}_{\mathcal{C}_{\mathcal{M}}}^{SI},\\
    &{\rm \textbf{v}}^{agent} = {\rm \textbf{W}}_{V}^{agent}\mathcal{F}_{A}^{SI}, {\rm \textbf{v}}^{map} = {\rm \textbf{W}}_{V}^{map}\mathcal{F}_{\mathcal{C}_{\mathcal{M}}}^{SI}.
\end{split}
\end{equation}
\begin{equation}
\begin{split}
    &{\rm \textbf{h}}^{A} = {\rm softmax}(\frac{\frac{1}{N_{a}+1}\sum_{i=0}^{N_{a}}{\rm \textbf{q}}^{agent}_i}{\sqrt{D}}\odot{\rm \textbf{M}}_{aff}){\rm \textbf{v}}^{map},\\
    &{\rm \textbf{h}}^{\mathcal{C}_{\mathcal{M}}} = {\rm softmax}(\frac{\frac{1}{N_{m}}\sum_{j=0}^{N_{m}-1}{\rm \textbf{q}}^{map}_j}{\sqrt{D}}\odot{\rm \textbf{M}}_{aff}^\mathsf{T}){\rm \textbf{v}}^{agent}.
\end{split}
\end{equation}
where ${\rm \textbf{W}}_{Q}^{agent}$, ${\rm \textbf{W}}_{Q}^{map}$, ${\rm \textbf{W}}_{V}^{agent}$, ${\rm \textbf{W}}_{V}^{map}$ are learnable linear projection matrices matrices for query and value.
The symbol $\odot$ is element-wise product while $i$ and $j$ denote the indexer of tensor. 
Afterwards, we obtain agent motion feature $\mathcal{F}_{A}^{Encoder}$ and coupled map feature $\mathcal{F}_{\mathcal{C}_{\mathcal{M}}}^{Encoder}$ after context fusion similar to Feed-Forward Network in Transformer \cite{vaswani2017attention}:
\begin{equation}
\begin{split}
    &\mathcal{F}_{A}^{Encoder} = {\rm MLP}({\rm LayerNorm}({\rm \textbf{h}}^{A}+\mathcal{F}_{A}^{SI})),\\
    &\mathcal{F}_{\mathcal{C}_{\mathcal{M}}}^{Encoder} = {\rm MLP}({\rm LayerNorm}({\rm \textbf{h}}^{\mathcal{C}_{\mathcal{M}}}+\mathcal{F}_{\mathcal{C}_{\mathcal{M}}}^{SI})).
\end{split}
\end{equation}

The bilateral query serves the same function as a 2-layer cross-attention, but with a more compact and parallel structure.
This is achieved by using a shared affinity matrix ${\rm \textbf{M}}_{aff}$, resulting in reduced time and space complexity.
Additionally, our module can be adapted to multi-head form similar to the original Transformer architecture.

\vspace{-0.6cm}
\section{Experiments}
\label{sec:exp}

\subsection{Experiment Setup}
\label{sub:bmk}

\noindent\textbf{Real-world Datasets.} We train and evaluate our model on three large-scale and challenging real-world datasets. 
Argoverse 1 \cite{chang2019argoverse} contains 333k 5-second sequences where each corresponds to a specific scenario sampled from a 290km-long roadway at 10 Hz. 
They provide trajectory histories, other agents and HD maps, with $(0,2]$ seconds for observation and $(2,5]$ seconds for prediction.
Argoverse 2 \cite{wilson2021argoverse} focuses on the prediction of multiple road users (vehicle, pedestrian, motorcyclist, cyclist and bus) with 250k 11-second scenarios sampled from 2110 km over six geographically diverse cities at 10 Hz. 
The dataset gives 5 seconds observation to predict the behaviors in the future 6 seconds.
nuScenes \cite{caesar2020nuscenes} is collected from 1000 scenes in Boston and Singapore, containing over 40000 samples, published at 2 Hz. 
The forecasting task for models is to predict the next 6 seconds according to the trajectory histories and HD maps in the past 2 seconds.

\noindent\textbf{Metrics.} We use the extensively adopted official metrics. 
1) minFDE$_{K}$: the minimum endpoint error between $K$ trajectories and ground-truth. 
2) minADE$_{K}$: the minimum average displacement error of each point between $K$ trajectories and ground-truth. 
3) MR$_{K}$: the proportion of scenarios where none of the predictions' endpoints are within 2.0 meters of ground-truth. 
4) brier-score: $(1.0$-$p)^2$ with $p$ as the probability of the trajectory with minimum endpoint error.
On Argoverse, we also report the official ranking metric brier-minFDE$_{K}$, indicating the minFDE$_{K}$ plus brier-score.

\noindent\textbf{Implementation Details.} We first divide the lane area within radius 50m from the interested agent into $N_{m}$=$128$ segments, which contains up to $P_{m}$=$31$ points with $d_{m}$=$15$ attributes. 
Within this area, we select $N_{a}$=$31$ agents closest to the interested agent. 
The relative motions in coupled map are defined as a $d_{r}$=$3$ vector $[dist,\cos\beta,\sin\beta]$ presenting distance ($dist$) and direction ($\beta$). 
All attention layers contain 4 heads and 128 hidden units. 
Notably, the quantity of learnable tokens and predicted trajectories $K$ is set 6 on Argoverse 1 and 2 \cite{chang2019argoverse, wilson2021argoverse} while 10 on nuScenes \cite{caesar2020nuscenes}.
Our model is trained for 200 epochs on 8 NVIDIA RTX 2080Ti GPUs. 
We use Adam optimizer with the initial learning rate of 1e-4, decaying to 1e-5 at 170$^{th}$ epoch and 1e-6 at 190$^{th}$ epoch, without weight decay.
We conduct experiments based on a small model with feature dimension $D$=$64$  and a large model with feature dimension $D$=$128$, termed as
MacFormer-S and MacFormer-L, respectively. 
The inference latency denotes the time required to predict 32 agents concurrently on a single NVIDIA RTX 2080Ti GPU.

\subsection{Ablation Study}
\label{sub:ablation}

We conduct ablation studies on the Argoverse 1 validation set using our 64-dimension MacFormer-S if not specified.

\vspace{-0.2cm}
\begin{table}[H]
    \scriptsize
    \begin{center}
        \caption{Importance of bilateral query on model performance.}
        \vspace{-0.2cm}
        \label{tab:bq_vs_stack}
        \renewcommand\arraystretch{1.1}
        \tabcolsep=0.6mm
        \centering
        \begin{tabular}{l|cccrr}
            \hline
            \specialrule{0em}{1pt}{1pt}
            Context fusion type & minFDE$_{6}$ & minADE$_{6}$ & MR$_{6}$ & Latency & $\#$Param \\
            \specialrule{0em}{1pt}{1pt}
            \hline
            \hline
            Bilateral query & \textbf{1.05} & \textbf{0.71} & \textbf{0.10} & \textbf{14ms} & \textbf{0.879M} \\
            Stack attention \cite{ngiam2022scene} & 1.08 & 0.73 & 0.11 & 83ms & 2.756M \\
            \hline
        \end{tabular}
        \vspace{-0.5cm}
    \end{center}
\end{table}

\noindent\textbf{Effectiveness of Bilateral Query.} 
By replacing bilateral query with stack attention \cite{ngiam2022scene} in our framework, we compare model performances of two context fusion manners. 
Specifically, we use 6 attention layers for stack attention (2 cross-attention and 4 self-attention) as the same as \cite{ngiam2022scene}.
As reported in Table.\ref{tab:bq_vs_stack}, bilateral query outperforms stack attention with 6x speed and using 68.1\% fewer parameters.
We validate the superiority of the proposed bilateral query in significantly reducing model parameters and inference latency. 

\begin{table}[H]
    \scriptsize
    \vspace{-0.2cm}
    \begin{center}
        \caption{Ablation studies on the proposed modules in our framework.}
        \label{tab:proposed_module_ab}
        \renewcommand\arraystretch{1.1}
        \tabcolsep=0.3mm
        \centering
        \begin{tabular}{c|c|c|c|c|ccc}
            \hline
            \multirow{3}*{$\;D\;$} & \multicolumn{2}{c|}{coupled map} & \multirow{3}*{\makecell[c]{$\,$bilateral$\,$ \\ query}} & \multirow{3}*{\makecell[c]{$\,$reference$\,$ \\ extractor}} & \multirow{3}*{minFDE$_{6}$} & \multirow{3}*{minADE$_{6}$} & \multirow{3}*{MR$_{6}$} \\
            \cline{2-3}
            & \begin{tabular}[c]{@{}c@{}}$\,$relative$\,$\\motions \end{tabular} & \begin{tabular}[c]{@{}c@{}}map\\$\,$topology$\,$ \end{tabular} & & & & \\
            \hline
            \hline
            64 & & \checkmark & \checkmark & \checkmark & 1.10 & 0.73 & 0.11 \\
            64 & \checkmark & \checkmark & & \checkmark & 1.15 & 0.75 & 0.13 \\
            64 & \checkmark & \checkmark & \checkmark & & 1.13 & 0.74 & 0.13 \\
            64 & \checkmark & \checkmark & \checkmark & \checkmark & 1.05 & 0.71 & 0.10 \\
            128 & \checkmark & \checkmark & \checkmark & \checkmark & \textbf{0.98} & \textbf{0.67} & \textbf{0.08} \\
            \hline
        \end{tabular}
    \vspace{-0.5cm}
    \end{center}
\end{table}

\begin{figure*}[t]
	\begin{center}
      \includegraphics[width=2.00\columnwidth]{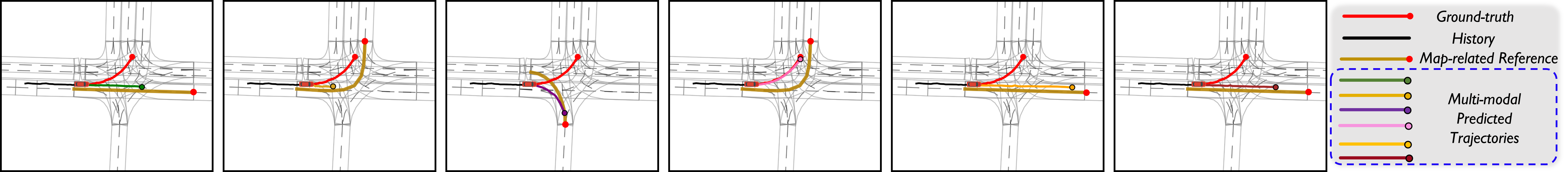}     
      \vspace{-0.5cm}
	\end{center}
   \caption{\label{fig:ref_vis} Qualitative visualization of map-related references produced by \textbf{reference extractor}. (Sect.\ref{sub:ablation})}
   \vspace{-0.6cm}
\end{figure*}

\noindent\textbf{Importance of Each Module.}
As shown in Table.\ref{tab:proposed_module_ab}, we illustrate the extent to which each module in our framework contributes to the prediction performance.
First, relative motions in coupled map establish explicit connection of map and trajectory, thereby enhancing the map utilization. 
Second, using bilateral query allows effective context fusion, facilitating accurate reference extraction from coupled map.
The absence of bilateral query results in a notable performance drop. 
Third, reference extractor has a significant impact on the performance, which provides map guidance for predictions. 
Without this module, the model cannot predict future trajectories using map-related references. 
The ablation of reference extractor is using agent motion feature to direct regress the predictions like implicit-fusion scheme instead of map-conditioned regression.
Moreover, we qualitatively visualize the most interested centerline of each modality according to the score matrix in cross-attention of reference extractor.
Fig.\ref{fig:ref_vis} shows each modality in final predictions is accurately guided by its corresponding reference.

\begin{table}[H]
    \scriptsize
    \vspace{-0.2cm}
    \begin{center}
        \caption{Ablation studies on multi-task optimization strategy (MTOS).}
        \label{tab:MTOS_ab}
        \renewcommand\arraystretch{1.1}
        \tabcolsep=0.6mm
        \centering
        \begin{tabular}{c|c|c|ccc}
            \hline
            \specialrule{0em}{1pt}{1pt}
            $\mathcal{L}_{primary}$ & $\mathcal{L}_{couple}$ & $\mathcal{L}_{capture}$ & minFDE$_{6}$ & minADE$_{6}$ & MR$_{6}$ \\
            \specialrule{0em}{1pt}{1pt}
            \hline
            \hline
            \checkmark &  &  & 1.14 & 0.77 & 0.13 \\
            \checkmark &  & \checkmark & 1.09 & 0.73 & 0.11 \\
            \checkmark & \checkmark &  & 1.11 & 0.74 & 0.10\\
            \checkmark & \checkmark & \checkmark & \textbf{1.05} & \textbf{0.71} & \textbf{0.10} \\
            \hline
        \end{tabular}
        \vspace{-0.5cm}
    \end{center}
\end{table}

\noindent\textbf{Ablation Studies on MTOS.} 
We evaluated the effect of each task in MTOS as presented in Table.\ref{tab:MTOS_ab}.
The coupled motion task $\mathcal{L}_{couple}$ can improve the performance since the map directly constrains predicted trajectories through optimizing future relative motions.
On the other hand, excluding the motion capture task $\mathcal{L}_{capture}$ leads to inferior performance, highlighting that this supervision plays a critical role in reinforcing the capture of motion priors from historical trajectory.

\vspace{-0.3cm}
\subsection{Results}
\label{sub:results}

\noindent\textbf{Comparison with State-of-the-art.} 
We compare our method with the state-of-the-art models on Argoverse 1$\&$2 and nuScenes benchmarks.
Gray region indicates the official ranking metric.
The single model results on Argoverse 1 is listed in Table.\ref{tab:av1_results}, where we also report the inference latency and model size using the official implementation if it exists.

\vspace{-0.1cm}
\begin{table}[H]
    \scriptsize
    \vspace{-0.2cm}
    \begin{center}
        \caption{Single model comparisons on Argoverse 1 benchmark\cite{chang2019argoverse}.}
        \label{tab:av1_results}
        \renewcommand\arraystretch{1.1}
        \tabcolsep=0.3mm
        \centering
        \begin{tabular}{l|>{\columncolor{gray!20}}c ccccrr}
            \hline
            Method  & \begin{tabular}[c]{@{}c@{}}brier- \\minFDE$_{6}$ \end{tabular} & \begin{tabular}[c]{@{}c@{}}brier- \\score \end{tabular} & minFDE$_{6}$ & minADE$_{6}$ & MR$_{6}$ & Latency & $\#$Param \\
            \hline
            \hline
            MacFormer-S & 1.9021 & 0.6317 & 1.2704 & 0.8490 & 0.1311 & \textbf{14ms} & \textbf{0.879M} \\
            MacFormer-L & \textbf{1.8275} & \textbf{0.6115} & 1.2160 & 0.8188 & \textbf{0.1205} & 19ms & 2.485M \\
            \hline
            HiVT \cite{zhou2022hivt} & 1.8422 & 0.6729 & \textbf{1.1693} & \textbf{0.7735} & 0.1267 & 69ms & 2.529M \\
            SceneTrans \cite{ngiam2022scene} & 1.8868 & 0.6547 & 1.2321 & 0.8026 & 0.1255 & 257ms & 15.296M \\
            TPCN \cite{ye2021tpcn} & 1.9286 & 0.6844 & 1.2442 & 0.8153 & 0.1333 & - & - \\
            DenseTNT \cite{gu2021densetnt} & 1.9759 & 0.6944 & 1.2815 & 0.8817 & 0.1258 & 531ms & 1.103M \\
            mmTrans \cite{liu2021multimodal} & 2.0328 & 0.6945 & 1.3383 & 0.8436 & 0.1540 & 129ms & 2.607M \\
            LaneGCN \cite{liang2020learning} & 2.0539 & 0.6917 & 1.3622 & 0.8703 & 0.1620 & 173ms & 3.701M \\
            \hline
        \end{tabular}
        \vspace{-0.5cm}
    \end{center}
\end{table}

Compared to these methods, MacFormer-S achieves the competitive or better performance using 64.5\% fewer parameters on average and with at least 10x lower latency.
MacFormer-L outperforms all other methods listed in Table.\ref{tab:av1_results} in terms of the official ranking metric brier-minFDE$_{6}$ and MR$_{6}$ with the lowest latency (more than 4x) and fewest parameters.
Also, MacFormer-L has the most reasonable probability prediction (best brier-score), which facilitates downstream decision-making and planning module for safe navigation. 
The above results demonstrate the superior prediction performance and real-time capability of our method.

\vspace{-0.1cm}
\begin{table}[H]
    \scriptsize
    \vspace{-0.2cm}
    \begin{center}
        \caption{Ensemble model comparisons on Argoverse 1 benchmark\cite{chang2019argoverse}.}
        \label{tab:av1_results_ensemble}
        \renewcommand\arraystretch{1.1}
        \tabcolsep=0.3mm
        \centering
        \begin{tabular}{l|>{\columncolor{gray!20}}c ccccrr}
            \hline
            Method  & \begin{tabular}[c]{@{}c@{}}brier- \\minFDE$_{6}$ \end{tabular} & \begin{tabular}[c]{@{}c@{}}brier- \\score \end{tabular} & minFDE$_{6}$ & minADE$_{6}$ & MR$_{6}$ & Latency & $N$ \\
            \hline
            \hline
            MacFormer-E & \textbf{1.7667} & \textbf{0.5526} & \textbf{1.2141} & 0.8121 & 0.1272 & \textbf{94ms} & 5 \\
            \hline
            MultiPath++ \cite{varadarajan2022multipath++} & 1.7932 & 0.5788 & 1.2144 & \textbf{0.7897} & 0.1324 & 738ms & 5 \\
            HO+GO \cite{gilles2021home,gilles2021gohome} & 1.8601 & 0.5682 & 1.2919 & 0.8904 & \textbf{0.0846} & - & - \\
            \hline
        \end{tabular}
        \vspace{-0.5cm}
    \end{center}
\end{table}

With the premise of ensuring the real-time requirement, we further improve the prediction performance using ensemble similar to \cite{varadarajan2022multipath++}.
The latency of ensemble model is calculated as $Nt_{s}+t_{p}$, where $N$ represents the number of single models, $t_{s}$ and $t_{p}$ are latency of single model and post-process respectively.
Our ensemble model, MacFormer-E, consists of 5 MacFormer-L with different random seeds. 
The results in Table.\ref{tab:av1_results_ensemble} show MacFormer-E outperforms other ensemble methods by a significant margin.

\vspace{-0.1cm}
\begin{table}[H]
    \scriptsize
    \vspace{-0.2cm}
    \begin{center}
        \caption{Comparisons on Argoverse 2 benchmark\cite{wilson2021argoverse}.}
        \label{tab:av2_results}
        \renewcommand\arraystretch{1.1}
        \tabcolsep=0.6mm
        \centering
        \begin{tabular}{l|>{\columncolor{gray!20}}c ccccr}
            \hline
            Method & brier-minFDE$_{6}$ & brier-score & minFDE$_{6}$ & minADE$_{6}$ & MR$_{6}$ \\
            \hline
            \hline
            MacFormer-E & \textbf{1.90} & \textbf{0.52} & 1.38 & \textbf{0.70} & 0.19 \\
            \hline
            GANet \cite{wang2022ganet} & 1.96 & 0.62 & \textbf{1.34} & 0.72 & 0.17 \\
            MTR \cite{shi2022motion} & 1.98 & 0.54 & 1.44 & 0.73 & \textbf{0.15} \\
            THOMAS \cite{gilles2021thomas} & 2.16 & 0.65 & 1.51 & 0.88 & 0.20 \\
            \hline
        \end{tabular}
        \vspace{-0.6cm}
    \end{center}
\end{table}

Comparisons between our best model with the state-of-the-art works on Argoverse 2 and nuScenes benchmarks is respectively listed in Table.\ref{tab:av2_results} and Table.\ref{tab:nus_results}.
The results show our method significantly outperforms these models.

\vspace{-0.1cm}
\begin{table}[H]
    \scriptsize
    \vspace{0.2cm}
    \begin{center}
        \caption{Comparisons on nuScenes benchmark\cite{caesar2020nuscenes}.}
        \label{tab:nus_results}
        \renewcommand\arraystretch{1.1}
        \tabcolsep=0.6mm
        \centering
        \begin{tabular}{l|>{\columncolor{gray!20}}c cccr}
            \hline
            Method & minADE$_{5}$ & minADE$_{10}$ & minFDE$_{1}$ \\
            \hline
            \hline
            MacFormer-L & \textbf{1.21} & \textbf{0.89} & 7.50 \\
            \hline
            PGP \cite{deo2022multimodal} & 1.27 & 0.94 & 7.17 \\
            THOMAS \cite{gilles2021thomas} & 1.33 & 1.04 & \textbf{6.71} \\
            GOHOME \cite{gilles2021gohome} & 1.42 & 1.15 & 6.99 \\
            \hline
        \end{tabular}
        \vspace{-0.6cm}
    \end{center}
\end{table}

\noindent\textbf{Discussion on the utilization of map constraints.}
Practically, there do exist abnormal driving behaviors violating map constraints.
If enforcing them as hard constraints, predicting abnormal behaviors is challenging.
Thus, our method integrates constraints via encouragement, enabling predictions satisfying constraints while allowing abnormal behavior forecasting.
Compared to constrained neural networks (enforcement manner), our method achieves more complete predicted behaviors and superior performance (Table.\ref{tab:constraint}).

\vspace{-0.2cm}
\begin{table}[h]
    \scriptsize
    \begin{center}
        \caption{Performance comparisons vs. constrained neural networks.}
		\vspace{-0.1cm}
        \label{tab:constraint}
        \renewcommand\arraystretch{1.1}
        \tabcolsep=1.0mm
        \centering
        \begin{tabular}{l|c|c|c|c}
            \hline
			\multirow{2}*{Method} & \multicolumn{2}{c|}{Argoverse 1 \cite{chang2019argoverse}} & \multicolumn{2}{c}{nuScenes\cite{caesar2020nuscenes}} \\
			\cline{2-5}
             & minFDE$_{6}$ & minADE$_{6}$ & minADE$_{5}$ & minADE$_{10}$ \\
            \hline
            \hline
            MacFormer-L & \textbf{1.22} & \textbf{0.82} & \textbf{1.21} & \textbf{0.89} \\
            \hline
            PRIME \cite{song2022learning} & 1.56 & 1.22 & - & - \\
            GoalNet \cite{zhang2021map} & - & - & 1.27 & 1.22 \\
            \hline
        \end{tabular}
        \vspace{-0.4cm}
    \end{center}
\end{table}

\begin{figure}[H]
	\begin{center}
      \includegraphics[width=0.85\columnwidth]{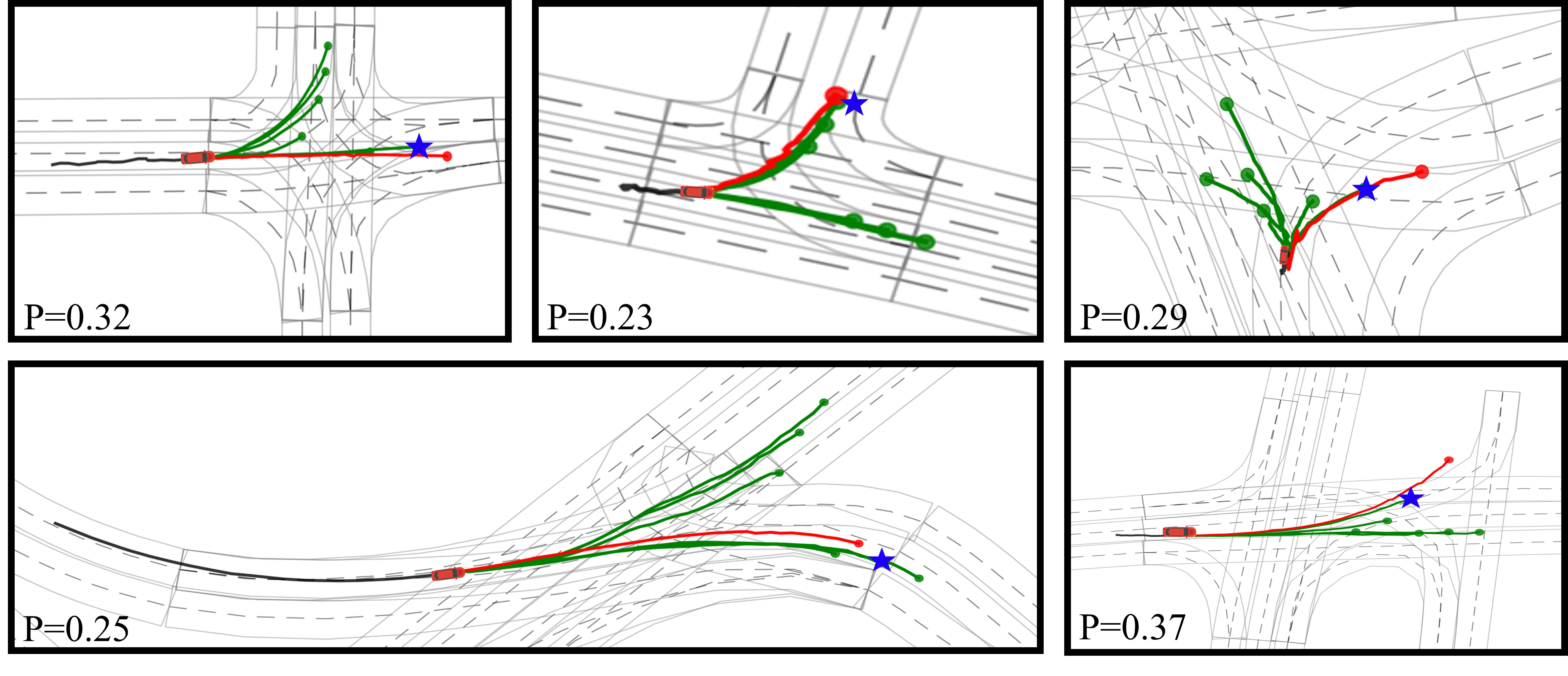}     
      \vspace{-0.4cm}
	\end{center}
   \caption{\label{fig:qualitative} Qualitative results of MacFormer-L. Historical trajectories is shown in \textcolor{black}{black}, ground-truth is shown in \textbf{\textcolor{red}{red}}, and predictions are shown in \textbf{\textcolor[RGB]{0,140,0}{green}}.
   \textbf{\textcolor{blue}{Blue}} star denotes the endpoint of the most confident predicted trajectory, where lower left corner shows its corresponding probability.
   }
   \vspace{-0.4cm}
\end{figure}

\noindent\textbf{Qualitative Results.} 
Fig.\ref{fig:qualitative} presents qualitative results of MacFormer-L on the Argoverse 1 validation set, including complicated intersections and long-range cases. 
Probability prediction results are also provided, with only the highest probability marked for clarity.
The results demonstrate our model can predict accurate and feasible trajectories with reasonable and plausible multi-modality satisfying map constraints in complex scenarios.
We further deployed our model in real-world scenarios using Argoverse tracking dataset \cite{chang2019argoverse} 
where we set $B$ as the number of all agents in the scenario for joint prediction.
The demonstration can be found at\footnote[1]{https://youtu.be/lCenh4XlH-4}.

\vspace{-0.2cm}

\begin{figure}[H]
	\begin{center}
      \includegraphics[width=0.8\columnwidth]{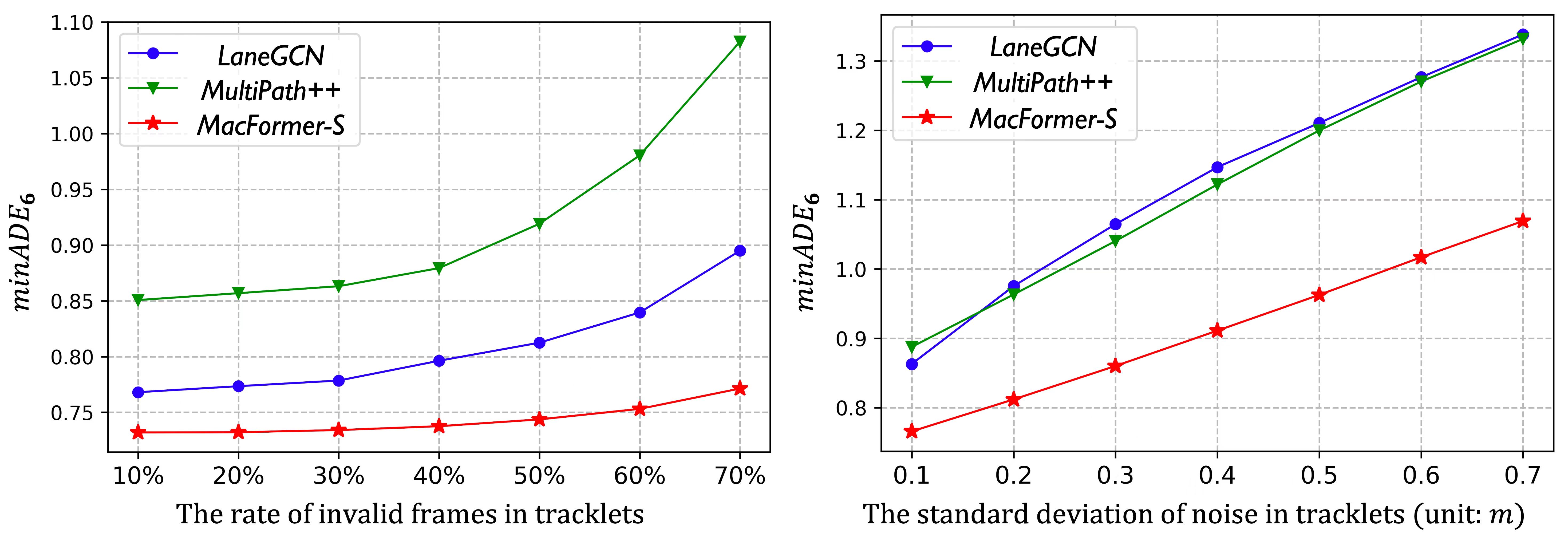}     
      \vspace{-0.5cm}
	\end{center}
   \caption{\label{fig:robust} The results of model robustness to imperfect tracklets.}
   \vspace{-0.1cm}
\end{figure}

\noindent\textbf{Robustness.}
Practically, history of agents sometimes contains invalid frames or noise, which poses a significant challenge to the robustness of trajectory prediction. 
We conclude three common cases: 1) new tracks with only a few frames, 2) missing frames in detection, and 3) noise in upstream processes.
To simulate these scenarios, we applied mask and random noise strategies to historical inputs and sent them to each prediction model on Argoverse 1 validation set.
All models used are the 64-dimension version.
Fig.\ref{fig:robust} shows that our method achieves superior performance compared to two state-of-the-art methods (LaneGCN \cite{liang2020learning} and MultiPath++ \cite{varadarajan2022multipath++}) as the rate of invalid frames or standard deviation of noise increases. 
Our approach exhibits minimal performance degradation while maintaining optimal robustness.

\vspace{-0.3cm}

\vspace{-0.2cm}
\noindent\textbf{Versatility.}
To showcase the versatility of our proposed framework, we conducted plugin-in experiments on Argoverse 1 validation set for classical models 
by directly replacing their map, context fusion, decoder, and output predictor with coupled map, bilateral query, reference extractor, and MTOS.
All models used are the 64-dimension version. 
The results in Table.\ref{tab:versa} demonstrate that our framework effectively improves prediction performance while reducing model parameters by 24.1\%, 56.2\%, and 23.7\% respectively for each component replaced.
These findings highlight the versatile capability of our framework for trajectory prediction.

\vspace{-0.2cm}
\begin{table}[H]
    \scriptsize
    \vspace{-0.2cm}
    \begin{center}
        \caption{The results of the versatility experiment on classical models.}
        \label{tab:versa}
        \renewcommand\arraystretch{1.1}
        \tabcolsep=0.6mm
        \centering
        \begin{tabular}{l|c|cccrr}
            \hline
            \specialrule{0em}{1pt}{1pt}
            Method & our framework & minFDE$_{6}$ & minADE$_{6}$ & MR$_{6}$ & $\#$Param \\
            \specialrule{0em}{1pt}{1pt}
            \hline
            \hline
            \multirow{2}{*}{LaneGCN \cite{liang2020learning}} &  & 1.17 & 0.76 & 0.13 & 1.749M \\
             & \checkmark & \textbf{1.09} & \textbf{0.73} & \textbf{0.10} & \textbf{1.326M} \\
            \hline
            \multirow{2}{*}{SceneTrans \cite{ngiam2022scene}} &  & 1.12 & 0.75 & 0.11 & 3.438M \\
             & \checkmark & \textbf{1.06} & \textbf{0.73} & \textbf{0.09} & \textbf{1.507M} \\
            \hline
            \multirow{2}{*}{Multiath++ \cite{varadarajan2022multipath++}} &  & 1.36 & 0.86 & 0.18 & 1.645M \\
             & \checkmark & \textbf{1.25} & \textbf{0.80} & \textbf{0.14} & \textbf{1.254M} \\
            \hline
        \end{tabular}
        \vspace{-0.5cm}
    \end{center}
\end{table}

\section{Conclusion}
\label{sec:conclusion}

We propose a real-time and robust trajectory prediction framework that effectively incorporates map constraints into the network. 
This is achieved by directly coupling map with trajectories and prediction conditioned on map-related references, respectively using coupled map and reference extractor.
To promote learning map constraints, we present a multi-task optimization strategy (MTOS).
Also, we develop a bilateral query scheme enhancing the computational efficiency for context fusion. 
Experiments show \textbf{MacFormer} achieves state-of-the-art performance on Argoverse 1$\&$2, and nuScenes real-world benchmarks.
Our model satisfies the requirements in terms of speed and model size, making it suitable for practical applications.
Results also provide strong evidence for the robustness and versatility of our framework.

\addtolength{\textheight}{-12cm}   








\bibliography{references}

\end{document}